\pdfoutput=1
\documentclass[letterpaper, 10 pt, conference]{ieeeconf}  %

\usepackage{subcaption}
\usepackage{graphicx}
\usepackage{color}
\usepackage[utf8]{inputenc}

\usepackage{amsmath}
\usepackage{amssymb}
\usepackage{url} 
\usepackage[section]{placeins}
\usepackage{multirow} 
\usepackage{tabularx}

\IEEEoverridecommandlockouts                              %

\title{\LARGE \bf
Energy-Score Stabilized Attentive Mixture-Density Policy: \\A Lightweight Approach to Multimodal Robot Motion
}

\author{
Sachiya Fujita$^{1}$, Hideyuki Ichiwara$^{1}$, Shigeki Sugano$^{1}$ and Tetsuya Ogata$^{1}$
\thanks{\raggedright $^{1}$All authors are with the Waseda University, Shinjuku, 169-8555 Tokyo, Japan. (Corresponding author: {\tt\small sachiya.fujita@akane.waseda.jp})}%
}

\begin{document}

\maketitle
\thispagestyle{empty}
\pagestyle{empty}

\begin{abstract}
For practical 
motion generation, it is essential to handle multimodality---multiple actions can be correct for the same observation---while maintaining low inference cost and strong generalization from limited demonstrations. We introduce an attentive mixture-density policy trained with the Energy Score that represents future visuomotor trajectories as explicit probability distributions with a single forward pass. Concretely, we output Gaussian mixture parameters for joint commands (and optionally attention points) and optimize them with the Energy Score 
(computed as a sum of 1D CRPS under a diagonal covariance assumption) 
instead of the negative log-likelihood (NLL) to avoid MDN singularities and gradient-scale instabilities. Our architecture couples explicit spatial attention with mixture-density outputs, enabling robust multimodal behavior selection under visual noise and small-data regimes. On an 8-DOF manipulator, the method generates distinct avoidance trajectories and stabilizes learning in obstacle avoidance. Moreover, in a randomized moving-object grasping task, where spatiotemporal multimodality and fast timing decisions are required, the proposed model outperforms Diffusion Policy by 15--25\% 
under several unseen conditions, demonstrating superior generalization and real-time feasibility.

\end{abstract}

\section{INTRODUCTION}
Recent advances in imitation learning (IL) have pushed robot skill acquisition beyond laboratory benchmarks toward deployment on real robots. A key missing capability, however, is robust manipulation of randomly moving targets in real time—tasks that require continuous tracking, rapid timing decisions, and safe execution under strict onboard compute budgets. 
Such capabilities are increasingly relevant for high-value applications in dynamic settings, including automated capture/restraint of laboratory animals, handling of moving specimens, and time-critical retrieval tasks, where environment dynamics and safety constraints make repeated re-planning or cloud computation impractical.

To enable these applications, IL methods must satisfy three requirements simultaneously: (i) low inference cost for resource-constrained robots, (ii) high generalization performance with limited demonstrations, and (iii) explicit handling of task multimodality, where multiple correct actions exist 
per state. Multimodality arises both spatially (e.g., choosing left vs. right paths around an obstacle) and temporally (e.g., selecting grasp timing for periodic or irregular target motion). However, existing approaches often satisfy at most two 
requirements. Deterministic regressors such as the Spatial Attention Point Network (SARNN)\cite{ichiwara22contact} achieve excellent data efficiency and lightweight inference but suffer from mode averaging in multimodal tasks\cite{florence22implicit}. Conversely, expressive probabilistic models such as diffusion-based policies\cite{chi25diffusion} capture multimodality but require iterative sampling and substantially larger datasets, which can be prohibitive for real-time onboard control.

We propose an energy-score stabilized attentive mixture-density policy that enables real-time multimodal motion generation and long-horizon tracking from small datasets. Our method outputs Gaussian mixture parameters for future joint trajectories (and optionally attention points), allowing multiple candidate behaviors to be represented explicitly as a probability distribution. 
Crucially, we replace Negative Log-Likelihood with the Energy Score 
(implemented as a sum of independent CRPS terms for computational efficiency)
\cite{gneiting07strictly}, 
evaluating 
the distance between the predicted distribution and ground truth without variance-collapse singularities, thereby stabilizing MDN training in multimodal settings. 
While our implementation leverages SARNN’s spatial attention mechanism for robust perception, the central contribution is a general and practical recipe: combining attentive state abstraction with energy-score–trained mixture-density 
outputs for stable, low-cost multimodal control.

To validate the method, we conduct two real-robot experiments with an 8-DOF manipulator: obstacle avoidance (spatial multimodality) and grasping a randomly moving object (spatiotemporal multimodality). The latter demonstrates that our policy can track and intercept unpredictable target motion under unseen patterns and speed changes—an ability directly aligned with real-world dynamic manipulation demands.

The primary contributions are:
\begin{enumerate}
    \item Energy-score stabilized Attentive MDN Policy: A lightweight policy framework that couples spatial attention with explicit mixture-density outputs for multimodal action generation.
    \item Stable Multimodal Learning: Energy Score training that avoids MDN singularities and improves convergence when multimodal distributions and auxiliary reconstruction losses coexist.
    \item Real-robot Dynamic Tracking: A randomized moving-object grasping benchmark on hardware, showing strong generalization under unseen trajectories and speeds with onboard real-time inference.
\end{enumerate}

\section{Related Work}
Robot behavior generation is broadly categorized into deterministic and probabilistic models. A representative deterministic model, 
SARNN
\cite{ichiwara22contact}, estimates spatial coordinates (attention points) 
for the task
from input images and learns temporal variations alongside joint angles using a Recurrent Neural Network (RNN). This approach enables efficient behavior learning by focusing on visually significant regions and exhibits high generalization performance against changes in object positions. Furthermore, the use of an RNN for the temporal model offers the advantages of extremely low inference costs and high generalization with limited data. However, deterministic regression suffers from an inherent "averaging problem" when handling multimodal data where trajectories branch. 
This 
stems from 
the $L_2$ loss (MSE), 
which converges to the conditional expectation $E[y|x]$, potentially generating physically inappropriate actions
, like
moving straight toward an obstacle
at a bifurcation point.

Generative 
models that treat output actions as probability distributions have been extensively studied. Urain et al. \cite{urain24deep} categorized methods for representing multimodality into five groups and compared their characteristics: (1) Diffusion, (2) Sampling, (3) Energy-based, (4) Categorical, and (5) Mixture Density Models.

(1) 
Diffusion models reconstruct distributions via iterative denoising
to offer 
expressive, non-parametric multimodal modeling. Diffusion Policy \cite{chi25diffusion} succeeds on complex tasks (e.g., Push-T), but iterative inference 
is costly for 
real-time dynamics; despite acceleration \cite{song20denoising,song23consistency}, 
few-shot generalization is limited for 
unseen speeds or placements.

(2) Sampling-based methods model complex distributions using latent variables. ACT \cite{zhao23learning} employs a CVAE to capture stochasticity and operator variability 
, enabling high-precision manipulation (e.g., opening Ziploc bags). However, its continuous Gaussian latent space can blur discrete modes (e.g., left vs. right avoidance), often producing unnatural intermediate actions. Similarly, ActionFlow \cite{funk24actionflow} applies Flow Matching \cite{lipman22flow} to joint-angle generation; however, due to FM’s strong linear constraint, it is prone to overfitting in low-data regimes, leading to degraded generalization \cite{zhu25diffusion}.

(3) Energy-Based Models (EBMs) \cite{lecun06tutorial} learn a function $E_\theta(a, s)$ 
assigning 
a scalar 
(energy) to a state $s$ and action $a$. A representative application in imitation learning is Implicit Behavioral Cloning (IBC) \cite{florence22implicit}. While EBMs 
allow 
highly flexible distribution representation, 
iterative optimization during inference
to find 
action 
$a$ that minimizes energy results in 
extremely 
high computational load, becoming a bottleneck in real-time robot control.

(4) Categorical models (e.g., Behavior Transformer (BeT) \cite{shafiullah22behavior}) treat multimodal action selection as a classification problem over discrete indices. However, their expressive power depends on 
the number of clusters, constraining flexible interpolation capabilities against environmental changes.

In contrast, (5) Mixture Density Models (MDNs) parameterize the output density as a Gaussian mixture (e.g., the behavior generation policy BC-RNN \cite{mandlekar21matters} with a GMM output layer \cite{bishop94mixture}), enabling low-cost one-pass inference. Compared to non-parametric diffusion models and discretized categorical approaches, MDNs provide a strong mixture-of-Gaussians inductive bias that is data-efficient and relatively hyperparameter-tolerant, since redundant components’ weights often shrink during training.

However, conventional BC-RNN structures directly input high-dimensional features processed by a CNN into the RNN without explicitly extracting object positions. Consequently, in small-data regimes, they are 
vulnerable to visual noise (e.g., background changes, object misalignments) 
leading to challenges in generalization. In contrast, SARNN \cite{ichiwara22contact} explicitly extracts "spatial coordinates to be focused on" as attention points through its spatial attention mechanism, making it robust against environmental changes even with limited data. Therefore, integrating SARNN's spatial recognition capabilities with the multimodal representation of MDNs is a rational approach to achieving advanced generalization with small datasets.

Nevertheless, while MDNs are structurally superior for representing multimodal action distributions, they face challenges in learning stability when using the Negative Log-Likelihood (NLL) loss. 
NLL is prone to numerical divergence (singularity problem) via variance minimization and causes scale and gradient mismatches with zero-bounded losses like MSE, complicating multimodal learning.
To address such instability, proper scoring rules---notably CRPS and its multivariate extension, the Energy Score (ES)---have been studied as objectives for probabilistic learning in other domains, including weather/forecasting (CRPS-based training) \cite{lang26aifs} and autonomous-driving trajectory distribution evaluation \cite{shahroudi24evaluation}. 
In robotics, there also exist works that incorporate energy-score-based objectives or energy-based formulations for multimodal behavior learning \cite{jia25fast,li25ebgan}. However, \emph{to the best of our knowledge}, prior robot motion-generation studies have not introduced ES/CRPS \emph{as a proper scoring-rule loss specifically to stabilize MDN (Gaussian-mixture) policies under multimodal sensory inputs}. 
Motivated by this gap, our research inherits the environmental robustness of SARNN's spatial attention mechanism while introducing the Energy Score (ES), a multivariate generalization of CRPS \cite{gneiting07strictly}, as a loss function to stabilize MDN training. In our implementation, we assume diagonal-independent covariances, so ES can be optimized as a sum of closed-form one-dimensional Gaussian CRPS terms (and their mixture extension) \cite{jordan19evaluating,grimit06continuous,hersbach00decomposition}. This design stabilizes variance convergence and fundamentally avoids the singularity problem. Furthermore, because it is based on the expectation of distances, it possesses high mathematical affinity with MSE, enabling stable convergence and high generalization in multimodal visuomotor policy learning that combines spatial attention with mixture density models.

\section{Proposed Method}
\begin{figure*}[htbp]
    \centering
    \includegraphics[width=\linewidth]{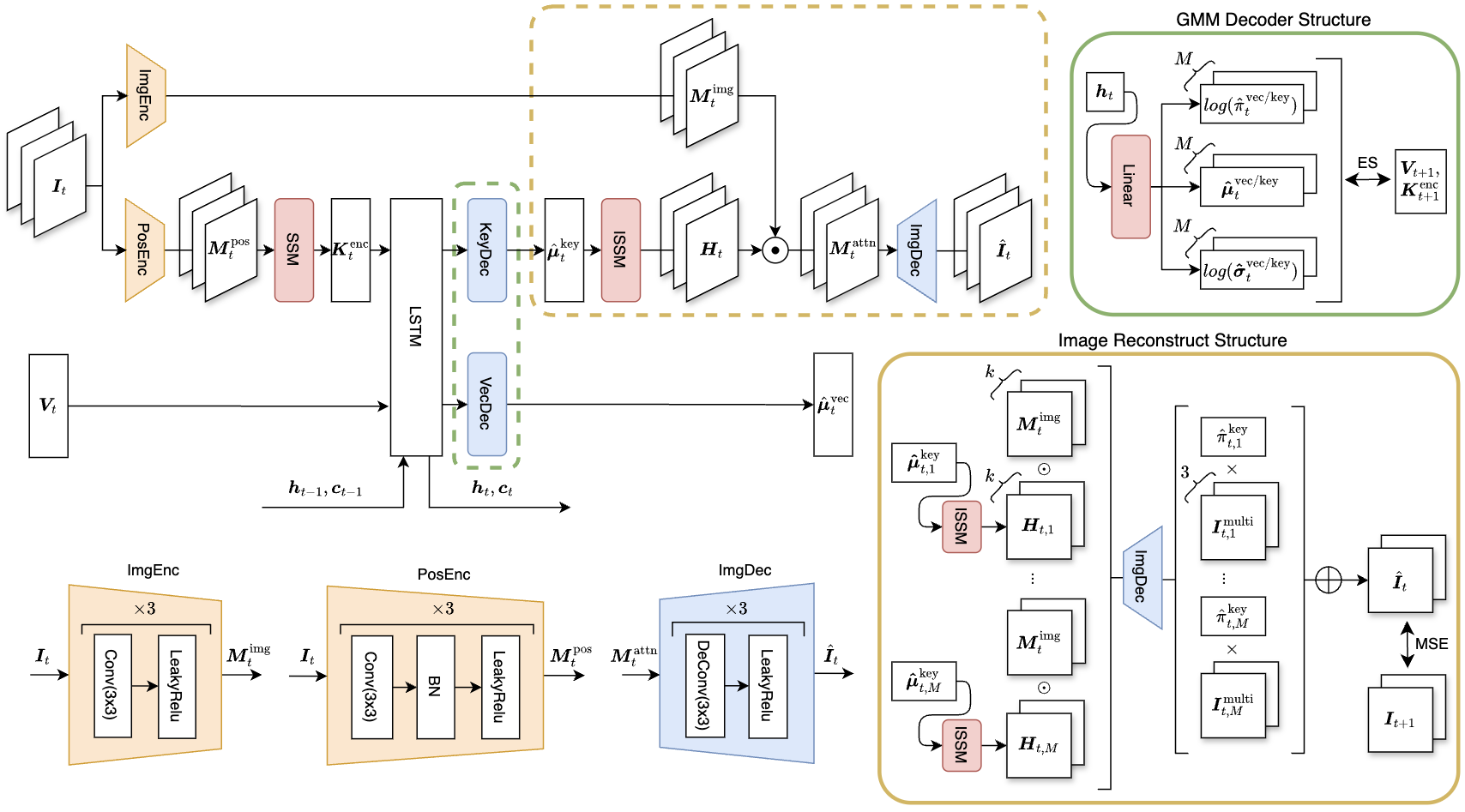}
    \vspace{-10pt}
    \caption{Architecture of the proposed model. The output layer (green box) represents the multimodality of future predictions by outputting joint angles and attention points as GMM parameters ($\pi, \boldsymbol{\mu}, \boldsymbol{\sigma}$). The Energy Score (CRPS) is adopted as the loss function. The yellow box indicates the reconstruction process optimized for the modified prediction format.}
    \label{fig:proposal_model_overall}
    \vspace{-10pt}
\end{figure*}

We propose an energy-score stabilized attentive mixture-density policy, termed GMMSARNN, that enables explicit multimodal action generation under real-time onboard constraints and small-data regimes.
The key idea is to predict future joint trajectories as a Gaussian mixture while maintaining stable training by optimizing the mixture parameters with the Energy Score (multivariate CRPS) instead of NLL.
We use an attentive recurrent encoder (SARNN-style) for robust perception; the idea is encoder-agnostic.

The overall architecture is illustrated in Fig. \ref{fig:proposal_model_overall}. The model integrates visual and robot state information to predict future states as probability distributions. First, $k$ attention point coordinates $\boldsymbol{K}^{\text{enc}}$ are extracted from the input image $\boldsymbol{I}$ via a CNN encoder and Spatial Softmax (SSM). These are concatenated with the current joint angles $\boldsymbol{V}$ to update the hidden state $\boldsymbol{h}$ using an LSTM. This configuration ensures robust feature extraction against background changes and effective integration of temporal information. From the hidden state $\boldsymbol{h}$, the GMM parameters—mixing coefficients $\pi$, means $\boldsymbol{\mu}$, and variances $\boldsymbol{\sigma}$—are computed via a linear layer as follows:
\begin{equation}
\begin{gathered}
\log\pi, \boldsymbol{\mu}, \log\boldsymbol{\sigma} = \text{Linear}(\boldsymbol{h}), \\
\pi = \text{softmax}(\log \pi), \quad \boldsymbol{\sigma} = \text{softplus}(\log\boldsymbol{\sigma})
\end{gathered}
\end{equation}
where Softmax is applied to $\pi$ to satisfy the summation constraint of 1, and Softplus is applied to $\boldsymbol{\sigma}$ to ensure positive values while maintaining numerical stability. To accommodate the $(-\infty, +\infty)$ range of $\boldsymbol{\mu}$, the network is configured to output $\log \pi$ and $\log \boldsymbol{\sigma}$.

The core design of our method lies in the extension of the output layer to a Mixture Density Network (MDN)(indicated by the green box in Fig. \ref{fig:proposal_model_overall}), which estimates predicted joint angles and attention points as a GMM with $M$ components. Depending on the target modality, we consider two configurations: "GMMSARNN-J," which models only joint angles as a distribution, and "GMMSARNN-JK," which models both joint angles and attention points. In this model, we adopt a structure where the distribution parameters ($\pi, \boldsymbol{\mu}, \boldsymbol{\sigma}$) for joint angles and attention points are estimated independently. This design is based on the distinct physical and visual characteristics of tasks. For instance, in an obstacle avoidance task, joint angles exhibit clear multimodality (e.g., "left avoidance" vs. "right avoidance") with different mixing coefficients $\pi$. Conversely, if the target of interest remains the same, the distribution of attention points remains unimodal regardless of the specific arm trajectory. Modeling these modalities independently ensures visual attention stability while allowing for flexible behavior selection.

Furthermore, we adopt the standard SARNN-style inverse spatial softmax (ISSM) to obtain mixture heatmaps $\boldsymbol{H}_{m} \in \mathbb{R}^{k \times H \times W}$ for $m=1, \dots, M$, which are element-wise multiplied by the attention map $\boldsymbol{M}^{\text{img}} \in \mathbb{R}^{k \times H \times W}$ to yield the attention-weighted feature maps $\boldsymbol{M}_{m}^{\text{attn}} = \boldsymbol{H}_{m} \odot \boldsymbol{M}^{\text{img}}$ (Fig. \ref{fig:proposal_model_overall}, yellow box). For efficiency, we batch the mixture dimension and run ImgDec once, producing the predicted images $\hat{\boldsymbol{I}}_m^{\text{multi}}$ for each component in a single forward pass. The final reconstruction is then formed by $\hat{\boldsymbol{I}} = \sum_{m=1}^{M} \pi_m \hat{\boldsymbol{I}}_m^{\text{multi}}$; this batch processing algorithm keeps inference real-time even as the number of components $M$ grows.Image reconstruction is used only as an auxiliary training loss to support the learning of visual features. The mixtures are trained with the Energy Score, which under our diagonal assumption reduces to a sum of closed-form Gaussian CRPS terms \cite{jordan19evaluating,grimit06continuous,hersbach00decomposition}.

To address the numerical instability common in MDN training, we introduce the Energy Score (ES) to directly evaluate the distribution shape. While a full-covariance ES captures inter-dimensional correlations, computing it requires computationally expensive operations such as $O(D^3)$ matrix inversions, which is prohibitive for our 10 Hz real-time control constraint. To achieve both theoretical stability and real-time inference, we assume independent diagonal covariance matrices for each of the $D$ dimensions of each modality. Although this restricts the explicit representation of intra-component correlations, the mixture of $M$ components can still implicitly capture complex multimodal joint distributions while significantly reducing the number of trainable parameters, thereby preventing overfitting in small-data regimes. Under this diagonal assumption, the ES mathematically decomposes into the analytically derived 1D Gaussian CRPS sum as the loss function:

\begin{equation}
\begin{split}
ES(F, y) = & \sum_{d=1}^{D} \left( \sum_{m=1}^{M}\pi_{m}A(y_d, \mu_{md}, \sigma_{md}) \right. \\
& \left. - \frac{1}{2}\sum_{m=1}^{M}\sum_{n=1}^{M}\pi_{m}\pi_{n}A(\mu_{md}, \mu_{nd}, \sqrt{\sigma_{md}^{2}+\sigma_{nd}^{2}}) \right),
\end{split}
\end{equation}
where the function $A(y, \mu, \sigma)$ represents the expected absolute deviation between an observation $y$ and a normal distribution $\mathcal{N}(\mu, \sigma^2)$, given in closed form as \cite{jordan19evaluating,hersbach00decomposition}:
\begin{equation}
A(y, \mu, \sigma) = 2\sigma \phi(z) + (y - \mu) (2\Phi(z) - 1),
\end{equation}
where $z = (y - \mu)/\sigma$, with $\phi$ and $\Phi$ denoting the standard normal PDF and CDF, respectively.
The total loss $L_{\text{total}}$ is defined by integrating the Energy Scores for joint angles ($L_{\text{vec}}$) and attention points ($L_{\text{key}}$) with the image reconstruction error $L_{\text{img}}$ (MSE):
\begin{equation}
L_{\text{total}} = w_{\text{img}} L_{\text{img}} + w_{\text{vec}} L_{\text{vec}} + w_{\text{key}} L_{\text{key}}
\end{equation}
Crucially, this CRPS-based objective fundamentally resolves the singularity problem commonly encountered when using Negative Log-Likelihood (NLL) in MDNs. Since NLL evaluates the point probability density function (PDF), it diverges to $-\infty$ when a component's variance approaches zero (variance collapse). In contrast, CRPS evaluates the integrated squared difference between Cumulative Distribution Functions (CDFs). Because CDFs are strictly bounded in $[0, 1]$, the expected $L_2$ distance remains finite even as $\sigma \to 0$. This theoretical property, along with its high scale affinity with the MSE loss for image reconstruction, ensures robust and stable multimodal learning.

\section{Experiments}
\begin{figure}[t] %
  \centering
  \includegraphics[width=\linewidth]{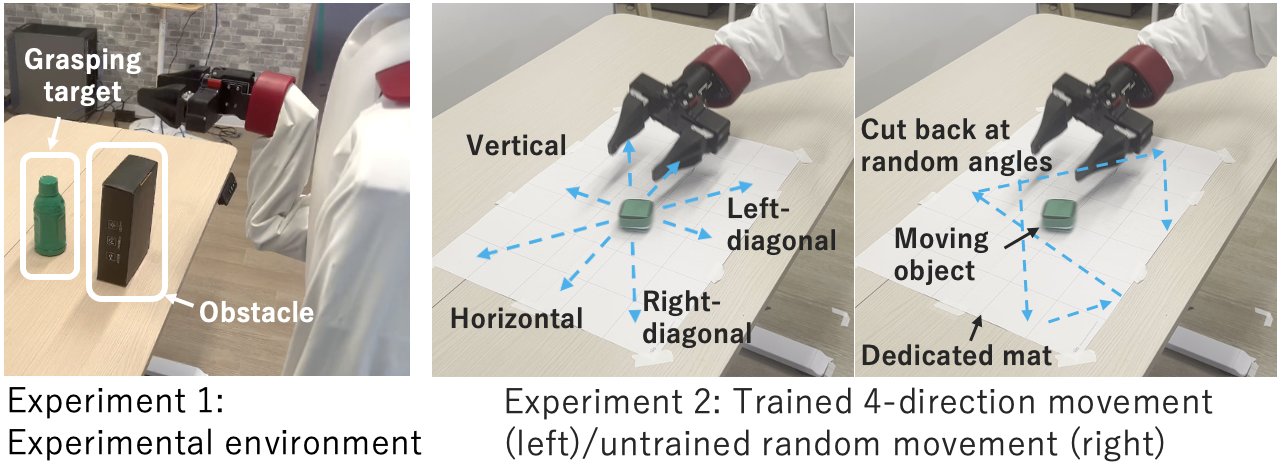} %
  \vspace{-10pt}
  \caption{Left: Experimental setup for Exp. 1. The target object is positioned behind an obstacle. Right: Movement patterns of the moving object in Exp. 2. During training, four reciprocating patterns are used: vertical, horizontal, right-diagonal, and left-diagonal. During inference, an additional unlearned pattern with random turn-back angles is introduced.}
  \label{fig:task_env} %
  \vspace{-10pt}
\end{figure}
To verify the effectiveness of the proposed method, we conduct a two-stage evaluation experiment using a real robot. This experiment focuses on how the system can adapt to complex multimodality while maintaining low inference costs and generalization performance with small datasets. In Experiment 1, to evaluate the capability to handle basic "spatial multimodality," the robot performs a task of grasping an object placed behind an obstacle by avoiding it from either the left or right side. Through this, we verify the ability to generate multiple trajectories by extending the output to a probability distribution, as well as the contribution to learning stability and success rates by changing the loss function from NLL to Energy Score.

In Experiment 2, we conduct a grasping task for a moving object that continuously moves back and forth to evaluate the adaptation capability in a complex environment where "temporal multimodality," which requires timing judgment, is mixed with spatial multimodality.

\subsection{Hardware}
For hardware, we use the humanoid robot 
"AIREC-Basic" \cite{ito26airec}, 
equipped with an 8-DOF arm, a 1-DOF parallel gripper at the end-effector, and a head camera. Training data is collected via teleoperation through a remote control device attached to the back of the robot. During collection, by directly using the joint angles from the operation side for training, the effects of mechanical deflection and backlash are eliminated, allowing the operator's intention to be learned directly. The data sampling rate and the control cycle during online inference are set to 10 Hz, and a 50 Hz linear interpolation is applied to the command values to achieve smooth motion on the real machine.

\subsection{Tasks}
\paragraph{Grasping a Stationary Object with Spatial Multimodality}
In Experiment 1, the robot performs a task of grasping a plastic bottle by avoiding an obstacle on the desk to either the left or right (Fig. \ref{fig:task_env} left). Five horizontal positions (idx 0--4) are set for the target object. Among them, three positions—the left end, center, and right end (idx 0, 2, 4)—are used as teaching positions. Training is conducted with a small dataset of 48 samples in total, consisting of 8 avoidance maneuvers from each of the two directions for each position. The remaining two positions are used to evaluate spatial generalization performance under unlearned conditions. As evaluation metrics, in addition to the success rate on the real robot, we define the generation of inappropriate trajectories that deviate from the target direction or collisions with the front of the obstacle as "mode averaging" and quantify its occurrence rate.

\paragraph{Grasping a Moving Object with Spatiotemporal Multimodality}
In Experiment 2, we conduct a dynamic task of grasping a small robot, "toio" \cite{sie26toio}, 
moving 
periodically back and forth (Fig. \ref{fig:task_env} right). As training data, we record grasping motions 
where the moving object reciprocates in 
a total of 
four directions: "vertical, horizontal, right-diagonal, and left-diagonal." 
In total, 
80 samples are collected, with 20 samples per direction. The toio travels on a dedicated mat\cite{switch26toio}, 
allowing 
control over its position and speed.

During data collection, specific measures are taken to incorporate temporal multimodality. First, the object is set in motion before the task starts, and the timing at which the operator begins the motion is randomized to diversify the object's position (phase). 
Second, by including data 
of successful trials 
after up to two retries, we record various motion patterns that attempt grasping at different timings.

During online inference, in addition to the four learned directions, we introduce "randomization of the movement trajectory" and "acceleration of movement speed (1.5x, 2.0x)" as more challenging unlearned conditions. In the unlearned patterns, every time the object reaches an edge, the next target is determined randomly based on the following Eq. (\ref{eq:task2_random_target}). This establishes an environment that demands high adaptation capabilities in both spatial and temporal dimensions, where $\theta$ is the object's angle relative to the mat, and $r$ is its distance from the center:
\begin{equation}
\begin{cases}
\theta_{\text{next}} = \theta_{\text{current}} + 180^\circ + \delta, & \delta \sim \mathcal{U}(-90^\circ, 90^\circ) \\
r_{\text{next}} \sim \mathcal{U}(13.6\,\mathrm{cm}, 19.6\,\mathrm{cm})
\end{cases}
\label{eq:task2_random_target}
\end{equation}

\subsection{Evaluation}
To quantitatively evaluate the effectiveness of the proposed method in terms of representation capability, learning stability, and generalization performance, we prepare several comparison models and ablation models. First, we selected Diffusion Policy (DP), which uses a diffusion model, and Action Chunking with Transformers (ACT), which is based on a Transformer, as representative stochastic models in current imitation learning. For DP, we adopted the most versatile CNN-based architecture, considering learning in a small-data environment. While ACT represents multimodality using CVAE, its nature of having ambiguous mode boundaries due to the continuity of the latent space led us to include "ACT with GMM," which explicitly applies a GMM to the output layer, in our comparison. This allows us to verify the difference between implicit multimodal representation by CVAE and explicit representation by GMM.

Next, to 
evaluate the contribution of the proposed method's components to the success rate, we construct 
ablation models. We use SARNN, a conventional deterministic model, as a baseline to confirm the impact of the "averaging problem" on multimodal data. We also prepare Stochastic SARNN (SCSARNN), which uses a unimodal Gaussian distribution ($M=1$) for the SARNN output, to verify the superiority of mixture distributions (GMM) over simple stochastic extension. Furthermore, we construct GMMSARNN-J$_{NLL}$, which uses Negative Log-Likelihood (NLL) instead of Energy Score as the loss function, to evaluate the impact of singularity problems and numerical instability in MDN training on motion generation in the real machine.

For MDN-based models, the number of GMM mixtures was 
$M=4$. This value was chosen to balance representation power and stability; in our experiments, increasing $M$ beyond this point caused some mixture components to become inactive—with mixing coefficients remaining near zero—effectively leading to mixture degeneracy.
For all SARNN-based models, the number of attention points $k$ extracted by the spatial attention mechanism is 
6. 
For ACT-based models, the Chunk Size, which is the number of action steps predicted at once, is
10. However, since Experiment 2 is a timing-sensitive task, inference is performed with Temporal Ensembling disabled to prevent smoothing from previous predictions from hindering agile motion.

Regarding the inference environment for each model, low-load models such as the proposed method, other SARNN-based models, and ACT-based models are executed using only the robot's built-in computational resources (NVIDIA Jetson Orin NX). On the other hand, for DP, which uses a diffusion model, it was difficult to maintain the 10 Hz control cycle essential for this task on the Orin NX due to its massive computational load. Therefore, an inference environment using an external PC (NVIDIA GeForce RTX 4090) is constructed only for DP. By explicitly separating the devices in this way, we verify the practicality of the proposed method for "on-board inference on edge devices," which is a primary focus of this study, while simultaneously clarifying the application limits of each model in real environments with severe computational resource constraints.

\section{Results}
In this section, we discuss the adaptation capability to multimodality, generalization performance, and characteristics of motion generation of the proposed method based on the results of two experiments using a real robot. 
Throughout this section, we emphasize \emph{practical} performance under the robot’s 10~Hz on-board control cycle (Jetson Orin NX) as the primary evaluation axis, and we report when external computation is required to meet the task’s timing constraints.

\subsection{Evaluation of Spatial Multimodality}
\begin{figure*}[t]
    \centering
    \includegraphics[width=\linewidth]{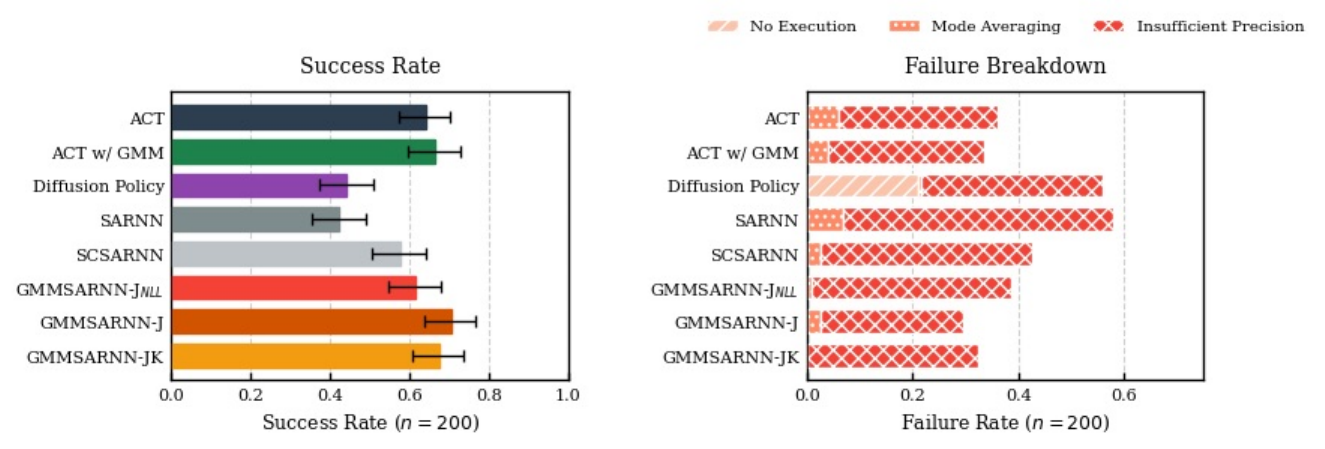}
    \vspace{-20pt}
    \caption{Comparison of Exp. 1 results. The left panel shows the success rate, and the right panel shows the failure rate and its breakdown. The left graph confirms that the proposed model achieves success rates equivalent to or higher than representative conventional models such as ACT and DP. In the right graph, the area enclosed by the crimson frame indicates the proportion of "mode averaging" generated; it can be seen that this proportion is significantly reduced in the proposed model compared to SARNN without GMM.}
    \label{fig:task1_result}
    \vspace{-10pt}
\end{figure*}

\begin{figure}[t]
    \centering
    \includegraphics[width=\linewidth]{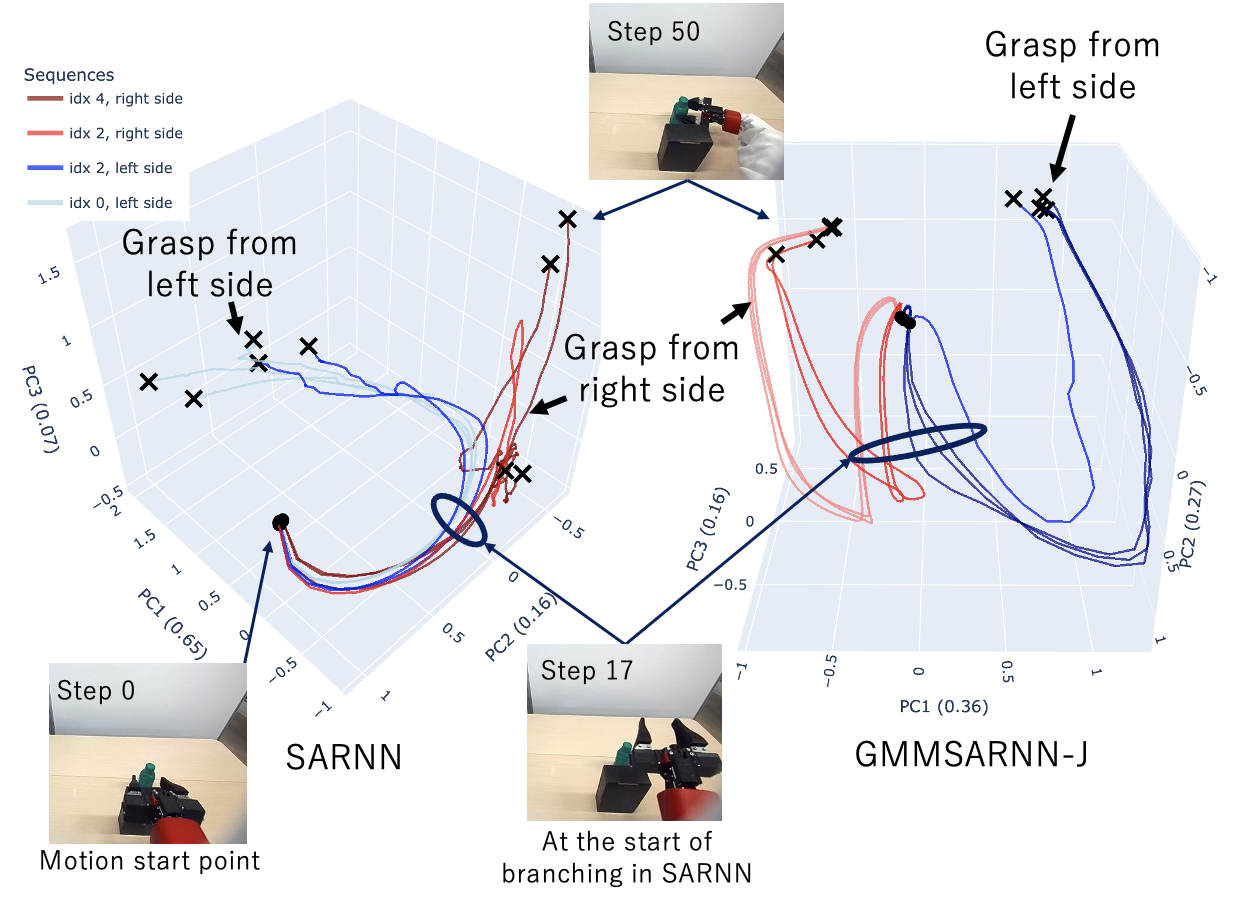}
    \vspace{-10pt}
    \caption{PCA visualization of internal states in Exp.~1 using only successful grasping trials. We plot nine trajectories in total: three trials each at the three seen target positions. }
    \label{fig:sarnn_based_model_pca}
    \vspace{-10pt}
\end{figure}

Fig. \ref{fig:task1_result} illustrates the success and failure rates, along with their breakdown, for each model in the obstacle avoidance task. In this experiment, a total of 200 trials (5 positions $\times$ 40 trials) were conducted for each model across five target positions, including unlearned ones.

As shown in the left panel of Fig. \ref{fig:task1_result}, the proposed models (GMMSARNN-J/JK) achieved a success rate of approximately 70$\%$, surpassing representative conventional methods such as ACT (approx. 65$\%$) and Diffusion Policy (DP, approx. 45$\%$). In particular, the proposed method outperformed DP with a significant performance margin, confirming its high adaptability to spatial multimodality tasks even in small-data environments. Furthermore, the success rate improved by 10--25$\%$ compared to the ablation models (SARNN, SCSARNN, and GMMSARNN-J$_{NLL}$), indicating that multimodal representation through GMM and learning stabilization via Energy Score (ES) significantly contribute to the performance improvement.

Focusing on the failure breakdown in the right panel of Fig. \ref{fig:task1_result}, the deterministic SARNN model experienced failures due to "mode averaging" in 7.0$\%$ of trials, whereas the proposed models significantly reduced this rate to 0--2.5$\%$. In SCSARNN, which utilizes a unimodal Gaussian distribution in the output layer, the mode averaging rate remained low at 2.5$\%$. However, since the success rates of the GMM-applied models (GMMSARNN-J/JK) were significantly higher, it is suggested that a simple unimodal distribution could not sufficiently accommodate subtle variations in target positions or compensate for a lack of precision. Similarly, in GMMSARNN-J$_{NLL}$ using NLL as the loss function, the mode averaging rate was as low as 1.5$\%$, comparable to the proposed method. However, the majority of its failures were attributed to "Insufficient Precision," resulting in a lower success rate than the proposed models using ES. This suggests that while NLL suffers from numerical instability during training, the adoption of ES (CRPS) allows the system to appropriately separate modes while maintaining high grasping precision.

The low success rate of Diffusion Policy was primarily due to a high frequency (20$\%$) of "No Execution," where the robot failed to initiate movement. This is likely because the small dataset was insufficient for DP to generalize motion patterns effectively. Consequently, the proposed method demonstrates superior learning efficiency and multimodality representation in small-data environments.

Furthermore, to investigate the process of acquiring multimodality, the transition of hidden states in the model's internal LSTM was visualized by compressing them into three dimensions using 
PCA 
for the motion from start to reaching (Fig. \ref{fig:sarnn_based_model_pca}). While the trajectories in the baseline SARNN overlapped until approximately step 17, the proposed method (GMMSARNN-J) showed clear branching into clusters corresponding to left and right avoidance paths at an early timing immediately after the start of the motion (indicated by black marks). This suggests that the introduction of GMM not only enhances the expressive power of the output layer but also has the effect of early separation and maintenance of each mode within the latent space.

\subsection{Evaluation of Temporal Multimodality}
\begin{figure*}[t]
    \centering
    \begin{minipage}[t]{0.49\linewidth}
        \centering
        \includegraphics[width=\linewidth]{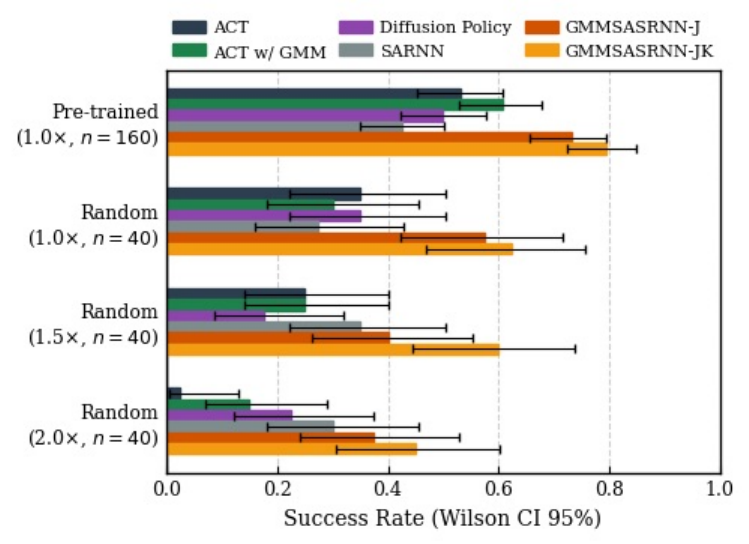}
        \caption{Success rates in Exp.~2: seen vs.\ unseen random patterns (1.0$\times$, 1.5$\times$, 2.0$\times$ speeds). Trials: 160 (seen), 40 per speed (unseen).}
        \label{fig:result_task2}
    \end{minipage}
    \hfill
    \begin{minipage}[t]{0.49\linewidth}
        \centering
        \includegraphics[width=\linewidth]{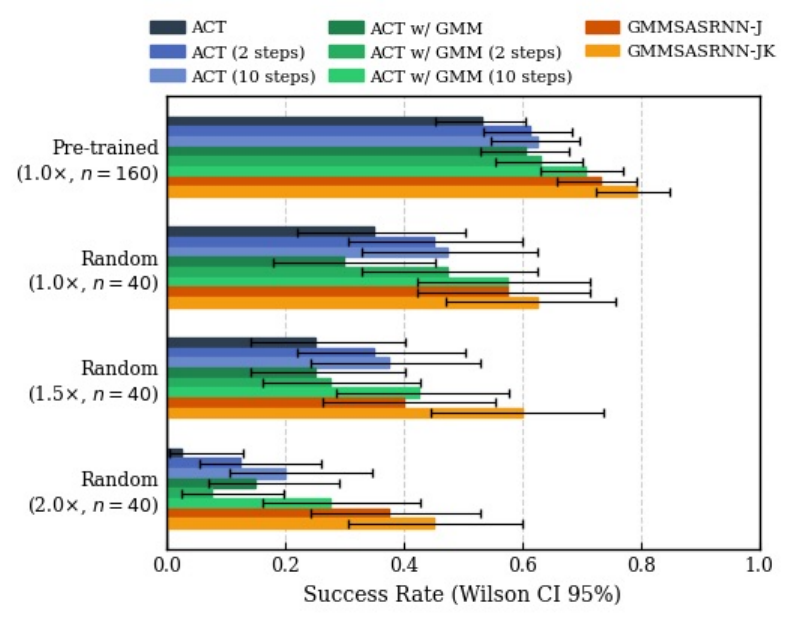}
        \caption{Comparison of success rates by input context length in Exp. 2. For ACT-based models with a 10-step context, inference was performed on an external PC because the inference speed on the robot's built-in PC fell below 10 Hz.}
        \label{fig:task2_context}
    \end{minipage}

    \vspace{-10pt}
\end{figure*}

\begin{table}[t]
    \centering
    \caption{Average inference times at Orin NX.}
    \label{tab:compare_inference_time}
    \small
    \newcolumntype{C}{>{\centering\arraybackslash}X}
    
    \begin{tabularx}{\columnwidth}{>{\hsize=1.4\hsize}C | >{\hsize=0.866\hsize}C >{\hsize=0.866\hsize}C >{\hsize=0.866\hsize}C} \hline \hline
        \multirow{2}{*}{Model} & \multicolumn{3}{c}{\textbf{Time [s]}} \\ \cline{2-4}
         & 1step & 2step & 10step \\ \hline
        ACT         & 0.0492 & 0.110 & 0.362 \\
        ACT w/ GMM  & 0.0578 & 0.0875 & 0.309 \\ \hline \hline
        
        Model & \multicolumn{3}{c}{\textbf{Time [s]}} \\ \hline
        SARNN       & \multicolumn{3}{c}{0.0397} \\
        GMMSARNN-J  & \multicolumn{3}{c}{0.0438} \\
        GMMSARNN-JK & \multicolumn{3}{c}{0.0626} \\ \hline
    \end{tabularx}
    \vspace{-10pt}
\end{table}

Fig. \ref{fig:result_task2} shows the success rates for the moving object grasping task. In this experiment, 160 trials were conducted for each model under the four pre-trained movement directions, and 40 trials were conducted for unlearned movement patterns.
Under pre-trained conditions, the proposed model improved the success rate by 15--25$\%$ compared to existing methods. Notably, the proposed method consistently outperformed the comparison models even under challenging unlearned conditions where movement speeds were increased to 1.5$\times$ and 2.0$\times$. Specifically, under the 2.0$\times$ speed condition, where ACT's performance almost collapsed and DP's success rate dropped to the 20$\%$ range, the proposed model (GMMSARNN-JK) maintained a success rate of 45$\%$, demonstrating high robustness against dynamic environmental changes.
The significant performance gap of the proposed model is attributed to its improved ability to track moving objects, which we hypothesize is supported by long-term context retention. 
In 
demonstration data, 
the operator finely adjusts the gripper to follow the object’s velocity and phase; reproducing this likely requires continuously predicting future reaching positions based on trajectories from the past few steps. Since the proposed model 
has 
a recurrent structure (LSTM), it can retain a context that covers the sequence length of this experiment (approximately 20 seconds), potentially allowing the model to accumulate periodic temporal information (e.g., the reciprocating motion) as its internal state. This enables the model to respond appropriately even to unlearned movement patterns as a “combination of learned patterns” or through interpolation.

In contrast, under our real-time settings, ACT and DP have limited access to past information. This restriction forces more reactive behavior that relies heavily on the current observation, 
leading to the observed performance gap.

While ACT typically leverages Temporal Ensembling for historical context, this smoothing introduces detrimental latency in dynamic tasks. Our preliminary tests showed enabling it reduced ACT’s success rate on pre-trained patterns from 53$\%$ to 45$\%$ (similarly degrading ACT with GMM).
Consequently, we intentionally disabled this feature to maximize ACT’s potential within the 10 Hz real-time constraint. However, this absence of context-carrying mechanisms necessitated predicting actions from only single-step observations. As a result, the model failed to identify the object's direction of movement and was limited to reactive behavior, responding solely to the target's current position at the current time.
Similarly, for Diffusion Policy (DP), the number of input observation frames is limited to the most recent two steps (0.2 seconds). For an object that switches directions on a several-second cycle, it is difficult to accurately estimate its phase and velocity with such a short window of past information, suggesting that it fell into reactive behavior similar to ACT. Indeed, in a comparative experiment (Fig. \ref{fig:task2_context}) where the sequence length of observation data input to ACT was expanded from 1 to 10 steps, the success rate improved as the context length increased, showing performance that approached that of the proposed model. This underlines that the accumulation of past information is critically important in tasks involving temporal multimodality.

However, as shown in Table \ref{tab:compare_inference_time}, ACT's average inference time increases with the number of input steps, reaching 0.362 s in the 10-step configuration—approximately seven times longer than the proposed model. This significantly exceeds the robot's 10 Hz control cycle, making real-time motion generation on edge devices difficult. In contrast, the proposed model achieves both low inference cost and high dynamic tracking performance by employing an RNN-based structure for efficient context retention, demonstrating clear superiority under practical computational resource constraints.

\subsection{Challenges of the Proposed Method}
During online inference in Experiment 2, instances of vibrations in the robot arm were observed. This phenomenon is likely attributed to rapid fluctuations in the mixing coefficients $\pi$ over short periods, caused by factors such as observation noise when performing the weighted synthesis of multiple outputs during inference. 
To address this phenomenon, we conducted experiments replacing the Softmax function with Sparsemax for the conversion of $\log \pi$ to $\pi$ to promote sparser and more stable mode selection. 
Unlike Softmax, which exhibits a "winner-takes-all" abrupt rise in $\pi$ due to its exponential nature, Sparsemax is piecewise-linear, suppressing explosive growth.
Experimental results confirmed that while this linear characteristic reduced arm vibrations by slowing transitions over the time axis, it simultaneously impaired the agility required to respond to temporal multimodality, ultimately leading to a decrease in the grasping success rate. Consequently, regarding the switching control of mixing coefficients in the proposed method, challenges remain in balancing stability during motion generation with responsiveness to temporal multimodality, necessitating further investigation.

\section{Conclusion}
In this study, we proposed “GMMSARNN,” a motion generation model that combines a lightweight attentive recurrent state encoder with an MDN output layer, balancing low inference costs with generalization performance on small datasets. In our implementation, we instantiate the encoder with an SARNN-style spatial attention mechanism.
By incorporating a Gaussian Mixture Model (GMM) into the output layer and stabilizing learning with Energy Score (multivariate CRPS), the proposed method overcomes the "averaging problem" inherent in conventional deterministic models. Simultaneously, the adoption of Energy Score resolves the singularity problem (numerical divergence) caused by Negative Log-Likelihood (NLL) in traditional Mixture Density Networks (MDNs), achieving stable trajectory learning.

Experimental results for the grasping task of a moving object—a spatially and temporally complex multimodal task—showed that the proposed model improved success rates by $15$--$20\%$ under pre-trained conditions compared to existing multimodal models. Furthermore, the proposed model consistently recorded higher success rates than comparison models even under unlearned conditions involving random movement patterns and speeds. Notably, while high-load models like Diffusion Policy require vast amounts of data, the proposed model demonstrated high generalization performance with limited data and proved capable of real-time on-board inference using only the robot's internal resources.

These results confirm that the proposed model is superior to existing methods under practical constraints of low computational resources and small datasets. This achievement accelerates the application of Imitation Learning (IL) to real-world tasks requiring high-level decision-making and rapid responsiveness in unstructured environments, such as the capture and restraint of laboratory animals. On the other hand, regarding the switching control of mixing coefficients in this method, challenges remain in balancing stability during motion generation with responsiveness to temporal multimodality, which warrants further investigation.

\end{document}